\pdfoutput=1

\documentclass[11pt]{article}

\usepackage[preprint]{acl}
\usepackage{times}
\usepackage{latexsym}

\usepackage[T1]{fontenc}

\usepackage[utf8]{inputenc}

\usepackage{microtype}

\usepackage{inconsolata}

\usepackage{graphicx}

\newcommand{\githublogo}{\protect\raisebox{-1pt}{\includegraphics[height=0.35cm]{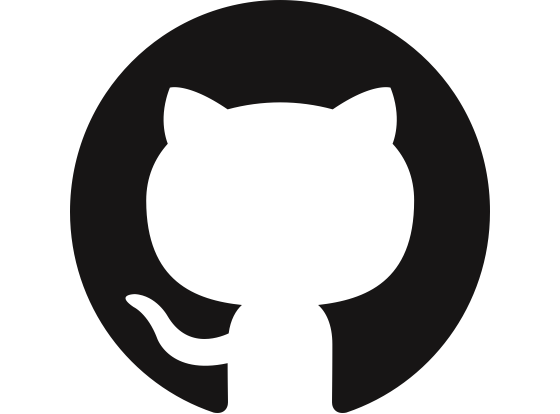}}}

\title{Trustworthy Agents for Electronic Health Records through Confidence Estimation}

\author{
    \textbf{Yongwoo Song\textsuperscript{1}},
    \textbf{Minbyul Jeong\textsuperscript{2}\thanks{Co-corresponding authors.}},
    \textbf{Mujeen Sung\textsuperscript{1}\footnotemark[1]}
    \\
    \small{
        \textsuperscript{1}Kyung Hee University,
        \textsuperscript{2}Upstage AI
    }
    \\
    \small{
        \texttt{
        syw5141@khu.ac.kr, minstar@upstage.ai, mujeensung@khu.ac.kr
        }
    }
}

\begin{document}
\maketitle

\begin{abstract}
Large language models (LLMs) have emerged as a promising approach for extracting and interpreting information from Electronic Health Records (EHR), offering possibilities for clinical decision support and patient care. However, the reliable deployment of LLMs in clinical settings remains challenging due to the risk of hallucinations. Thus, we propose Hallucination-Controlled Accuracy at k\% (HCAcc@k\%), a novel metric quantifying the accuracy-reliability trade-off at varying confidence thresholds. We also introduce TrustEHRAgent, a confidence-aware agent that incorporates step-wise confidence estimation for clinical question answering. Experiments on MIMIC-III and eICU datasets demonstrate that TrustEHRAgent outperforms baselines under strict reliability constraints, achieving substantial improvements of 44.23\%p and 25.34\%p at HCAcc@70\% while baseline methods fail completely at these stringent thresholds. These performance differentials under reliability constraints underscore the limitations of traditional accuracy metrics in evaluating healthcare AI agents. Our work contributes to the development of trustworthy clinical agents capable of delivering accurate information—or transparently expressing uncertainty—when confidence is low.\footnote{\href{https://github.com/ldi-kyunghee/TrustEHRAgent}{\githublogo\hspace{0.1cm}ldi-kyunghee/TrustEHRAgent}}
\end{abstract}


\begin{figure}[t]
    \centering
    \includegraphics[width=0.6\textwidth]{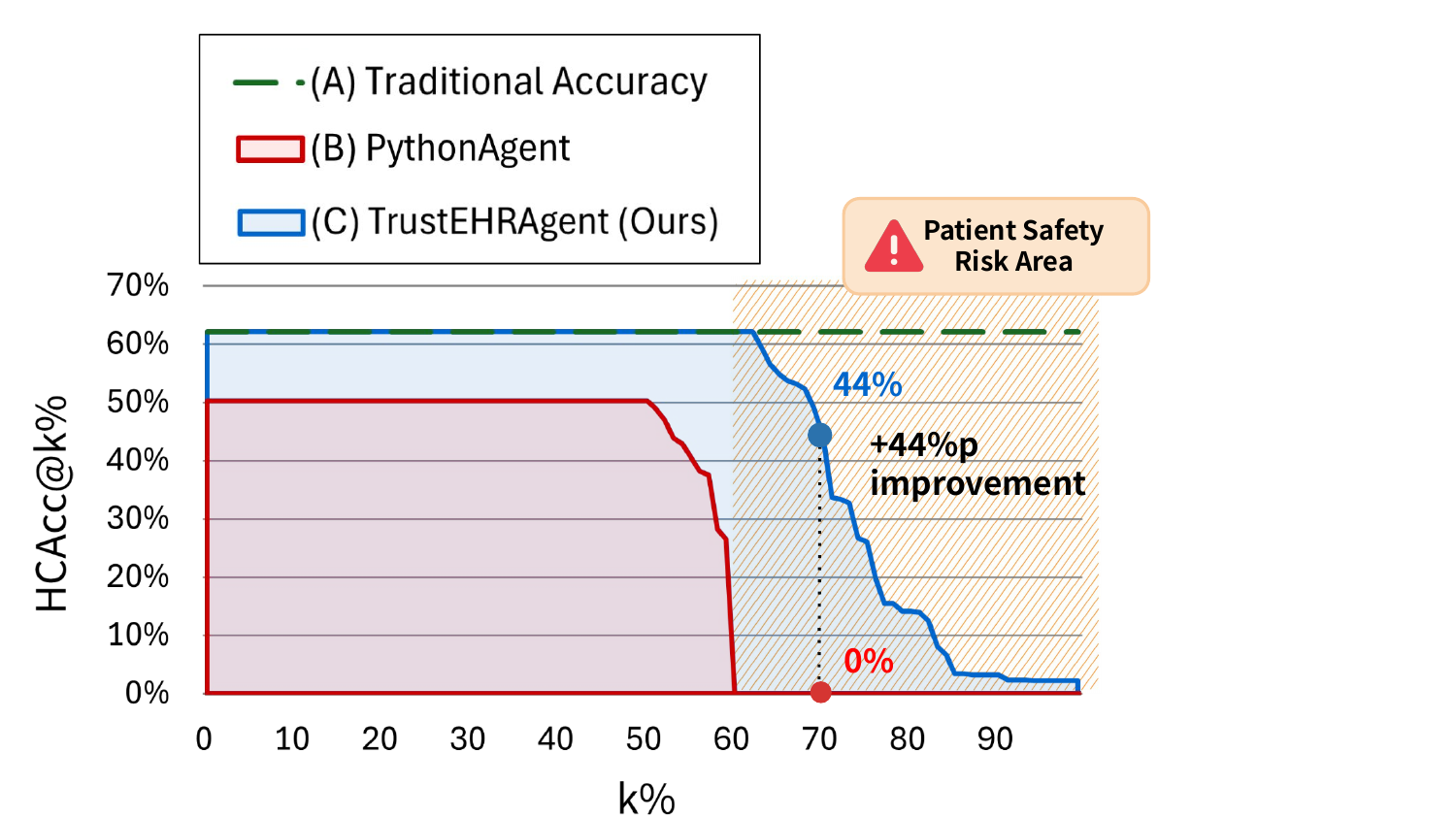}
    \caption{\textbf{Performance comparison under reliability constraints using the HCAcc@k\% metric, where k\% represents the minimum required reliability.} Traditional accuracy metrics (A) cannot identify the Patient Safety Risk Area, a region where strict reliability is essential for patient safety. In contrast, while (B) PythonAgent completely fails at $k \geq 60\%$, (C) Our TrustEHRAgent maintains stable performance under high reliability requirements, demonstrating the need for reliability-aware evaluation in medical AI agents.}
    \label{fig:hcacc_comparison}
\end{figure}

\section{Introduction}

Large language models (LLMs) have shown remarkable potential in healthcare, particularly in extracting and interpreting information from Electronic Health Records (EHRs) \cite{qiu2024llm}.
However, deploying these agents in clinical settings presents unique challenges, as hallucinations or factual errors can have serious implications for patient safety and care quality \cite{pal2023med, li2023halueval, ahmad2023creating}.
While prior works have largely focused on improving the accuracy of LLM-based clinical assistants \cite{wu2023medical}, comparatively little attention has been given to reliability—the model’s ability to recognize its own limitations and abstain from answering when uncertain.
This lack of reliability poses a critical barrier to the safe and responsible deployment of LLMs in real-world clinical environments \cite{hager2024evaluation}.

The widespread use of traditional evaluation metrics fails to capture the risks posed by hallucinations in high-stakes medical contexts (see Figure~\ref{fig:hcacc_comparison}A).
These metrics typically assume that agents attempt to answer \textit{every} question, ignoring the practical necessity of abstention in uncertain or ambiguous cases.
As shown in Figure~\ref{fig:hcacc_comparison}B, existing methods suffer substantial performance degradation under strict reliability constraints, underscoring their vulnerability in real clinical scenarios.
In clinical applications of AI, the ability to estimate confidence and recognize uncertainty is not merely desirable—it is essential. 
A reliable LLM-based agent must know when to respond and when to abstain, rather than risk producing harmful outputs in uncertain situations.

To address these limitations, we introduce Hallucination-Controlled Accuracy at k\% (HCAcc@k\%), a novel evaluation metric that quantifies the trade-off between accuracy and reliability across varying confidence thresholds.
Unlike traditional metrics, HCAcc@k\% explicitly incorporates reliability, making it well-suited for evaluating agents intended for clinical use.

Based on this evaluation metric, we propose TrustEHRAgent, a confidence-aware clinical agent that integrates step-wise confidence estimation into its question-answering environments.
This enables the agent to make informed decisions about whether to answer or abstain based on its internal uncertainty estimates.
As illustrated in Figure~\ref{fig:hcacc_comparison}C, our method demonstrates improved performance at stringent thresholds compared to existing approaches, providing a foundation for safer clinical decision support.
Notably, under stringent reliability constraints (HCAcc@70\%), TrustEHRAgent maintains performance with 44.23\% and 25.34\% accuracy on MIMIC-III and eICU datasets respectively while baseline methods achieve 0\% performance.
TrustEHRAgent also exhibits superior performance under traditional evaluation, achieving 62.13\% and 59.82\% accuracy (HCAcc@0\%), representing improvements of 3.16\%p and 6.72\%p over the existing methods.
Additionally, the agent shows the ability to appropriately abstain from answering when encountering ambiguous medical abbreviations, preventing potential errors through selective response mechanisms.

Our main contributions in this work include:
\begin{itemize}
   \item A novel evaluation metric (HCAcc@k\%) that quantifies the accuracy-reliability trade-off at varying confidence thresholds, enabling principled assessment of medical AI agents under patient safety constraints.
   
   \item A confidence-aware clinical agent framework (TrustEHRAgent) that achieves remarkable performance improvements of 44.23\%p and 25.34\%p at stringent reliability thresholds where existing methods fail completely.
   
   \item Evidence that traditional accuracy metrics inadequately reflect agent reliability in clinical environments, with methods showing similar baseline accuracy exhibiting dramatically different performance under safety constraints.

\end{itemize}
    


\begin{figure*}[t]
    \centering
    \includegraphics[width=1\textwidth]{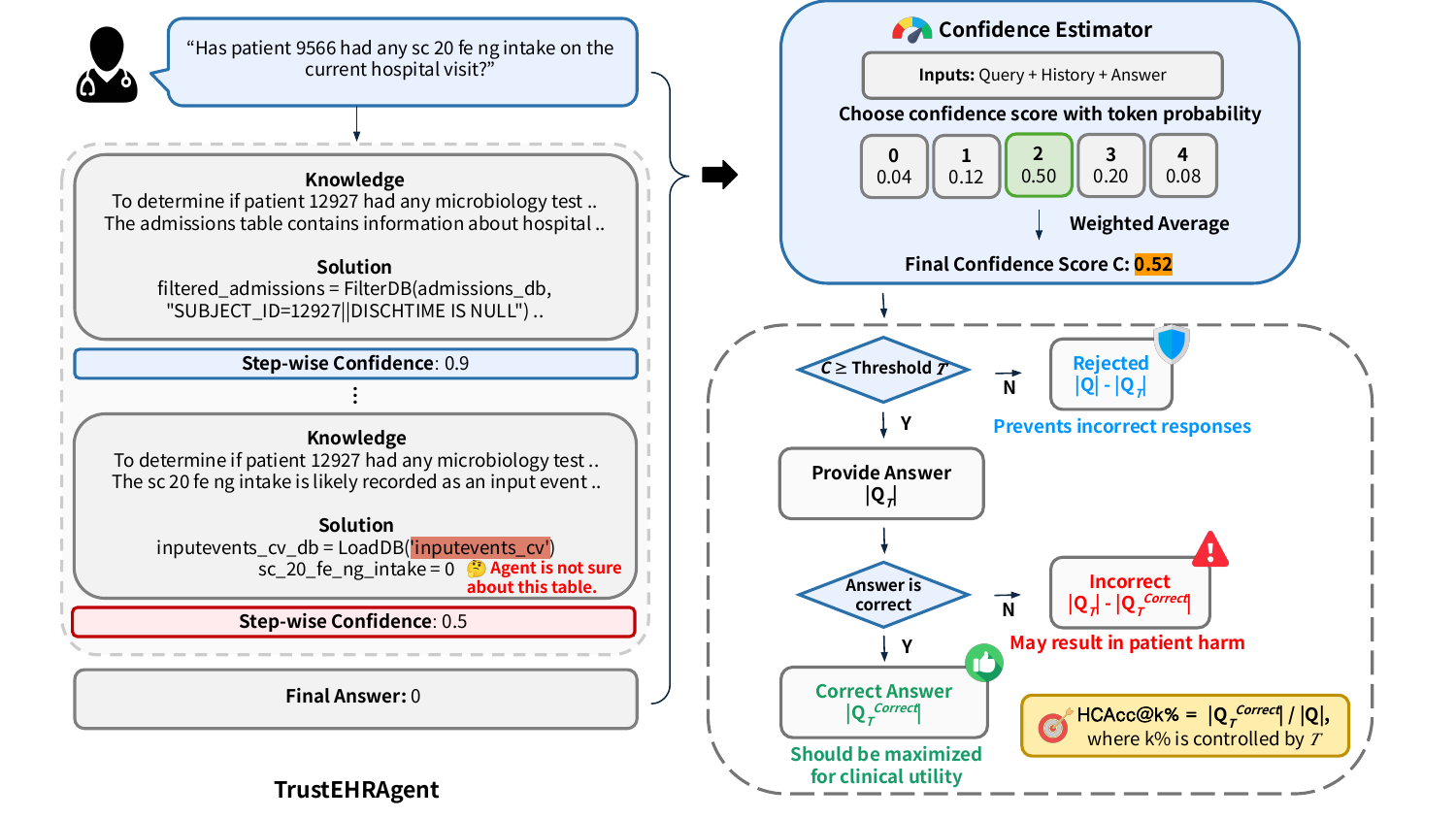}
    \caption{Overview of TrustEHRAgent framework. The left panel shows step-wise confidence estimation during clinical query resolution, where the agent assigns low confidence (0.5) when using uncertain database tables. The right panel illustrates the confidence estimator that takes the query, reasoning history, and answer as inputs to compute a final confidence score (C = 0.52) using token probabilities. The computed confidence score is then compared against a predefined threshold to make threshold-based decisions that either provide or reject answers based on reliability requirements. To achieve high HCAcc@k\%, the agent must provide accurate answers when confident and appropriately abstain when uncertain.}
    \label{fig:trustehragent_framework}
\end{figure*}

\section{Related Work}

\textbf{Reliability in Clinical LLMs}
Recent advances in LLMs have enabled their application across various medical domains \cite{wu2023medical, qiu2024llm}. For Electronic Health Record (EHR) systems\cite{wang2020text, yang2022large} introduced Text-to-SQL benchmarks for database querying, while \cite{shi2024ehragent} developed EHRAgent, an agent framework optimized for EHR interaction. Despite these advances, deployment in clinical settings faces significant challenges due to patient safety concerns, as factual errors can lead to harmful clinical decisions. To address these reliability challenges, several approaches have been proposed, including evaluation frameworks for medical question answering \cite{li2023halueval, agarwal2024medhalu, hager2024evaluation}, hallucination detection methods across NLP tasks \cite{ahmad2023creating, ahsan2024retrieving, mehenni2025medhal}, and medical-specific benchmarks for evaluating reasoning and memory-related hallucinations \cite{pal2023med}. However, these approaches lack integrated evaluation metrics that simultaneously consider both accuracy and reliability, particularly in clinical agent environments that require multiple reasoning steps.  

\textbf{Uncertainty and Confidence Estimation}
Confidence estimation for LLM outputs has become increasingly important for ensuring AI reliability in high-stakes applications \cite{mahaut2024factual}. Current approaches can be categorized into white-box methods that utilize internal model representations \cite{zhengbao2020, orgad2024llms, gekhman2025inside} and black-box methods employing sampling-based consistency measures and verbalized confidence techniques \cite{jiuhai2023, xiong2023can, lin2023generating, yang2024verbalizedconfidencescoresllms}. While recent research has emphasized evaluation methodologies beyond traditional accuracy metrics \cite{wang2023assessing, lin2021truthfulqa}, existing confidence estimation approaches primarily focus on single input-output relationships rather than evaluating reliability across multiple reasoning steps \cite{han2024towards}. This limitation is particularly problematic in clinical agent environments where errors can cascade through multiple reasoning steps, making it essential to assess confidence at each stage of the decision process.  
Our approach addresses these limitations by introducing HCAcc@k\%, a novel metric that quantifies accuracy-reliability trade-offs in medical domains, and TrustEHRAgent, a confidence-aware framework that integrates step-wise uncertainty estimation to enable safer healthcare AI agents.

\section{Methodology}
\label{sec:method}

\subsection{Problem Formulation}
We formulate reliable clinical question answering as a function $f: (q, D) \rightarrow (a, C)$. 
Given a clinical question $q$ about patient records and an EHR database $D$, the goal is to determine both an answer $a$ and a confidence score $C \in [0, 1]$ indicating the agent's certainty in the answer's correctness.
Importantly, the agent should be able to reject a query when confidence falls below a predetermined threshold $\tau$.
The confidence score is compared against the threshold $\tau$ to determine whether to provide an answer or reject the query.
\begin{equation}
\begin{cases}
\text{provide answer } a, & \text{if } C \geq \tau \\
\text{reject}, & \text{if } C < \tau
\end{cases}
\end{equation}

A key challenge lies in designing an agent that can accurately estimate its confidence during database interactions, establishing a threshold $\tau$ that optimally balances two competing objectives: maximizing the number of correctly answered queries while ensuring that the proportion of incorrect responses remains below an acceptable limit for clinical safety.

\subsection{TrustEHRAgent: A Confidence-aware Framework}
\label{sec:trustehragent}

We propose TrustEHRAgent, a framework that enhances reliability in clinical question answering environments. Figure~\ref{fig:trustehragent_framework} provides an overview of our framework architecture. Our approach builds upon the code-based reasoning capabilities of EHRAgent \cite{shi2024ehragent} while introducing two key components for improved reliability assessment: step-wise confidence estimation to track uncertainty throughout the reasoning process, and a dedicated confidence estimator that synthesizes these signals to compute the final confidence score $C$.

\textbf{Step-wise Confidence Estimation} 
Our framework employs a step-wise verbalization approach \cite{yang2024verbalizedconfidencescoresllms} to enhance reliability assessment throughout the reasoning process. As the agent works through each stage of clinical query resolution, it explicitly assigns confidence scores on a 0-10 scale at each reasoning step through prompt-based instructions. These scores reflect the data quality, query complexity, and interpretation ambiguity specific to each step, generating step-wise signals that serve as crucial inputs for our confidence estimation mechanism.
This approach enables the framework to track how uncertainty evolves throughout the reasoning process and identify potential points of failure before they propagate to the final answer. Implementation details are provided in Appendix~\ref{app:step-wise}.

\textbf{Confidence Estimator} 
Building upon the step-wise confidence signals, TrustEHRAgent incorporates a dedicated confidence estimation component that evaluates the overall trustworthiness of generated responses. This component analyzes three key inputs: the original clinical question, the agent's reasoning history enriched with step-wise verbalized confidence signals, and the proposed final answer.

The Confidence Estimator leverages a LLMs to evaluate response reliability by representing confidence as $N(=5)$ discrete levels ($S \in \{0, 1, ..., S_{\max}\}$). To achieve more fine-grained scoring and mitigate scoring bias issues, we utilize token log-probabilities rather than direct discrete outputs \cite{liu2023g}. We derive a probability distribution over these levels using $P(S=i) = \exp(\text{logprob}_i)$ and compute a weighted average as shown in Equation~\ref{eq:confidence}.

\begin{equation}
C = \frac{\sum_{i=0}^{S_{\max}} i \cdot P(S=i)}{S_{\max}}
\label{eq:confidence}
\end{equation}

We then normalize this weighted score by $S_{\max}$ to obtain the final confidence score $C \in [0,1]$. Implementation details are provided in Appendix~\ref{app:confidence}.

\subsection{Threshold-based Decision Making}
The confidence score $C$ is then compared against a predefined reliability threshold $\tau$ to make clinical decision support determinations.

When $C \geq \tau$, the agent provides the answer, which is then categorized as either Correct Answers ($Q_\tau^{\text{Correct}}$) or Incorrect Answers ($|Q_\tau| - |Q_\tau^{\text{Correct}}|$). When $C < \tau$, the agent rejects the query to prevent potential clinical errors ($|Q| - |Q_\tau|$).

The threshold $\tau$ serves as a hyperparameter that controls the trade-off between agent utility and patient safety. Higher thresholds prioritize safety by only providing highly confident responses, while lower thresholds allow broader utility with reasonable reliability standards. This enables healthcare providers to adapt the agent's behavior based on clinical context and risk tolerance.

Through this comprehensive framework, TrustEHRAgent optimizes the balance between accuracy and reliability by providing accurate answers when confident and appropriately abstaining when uncertain, ensuring safe clinical AI support without compromising patient safety.

\subsection{Hallucination-Controlled Accuracy at k\% (HCAcc@k\%)}
\label{sec:hcacc}

To evaluate the effectiveness of our confidence-aware framework, we introduce Hallucination-Controlled Accuracy at k\% (HCAcc@k\%). Traditional accuracy metrics inadequately evaluate clinical AI agents as they assume models should attempt every question regardless of uncertainty. In high-stakes medical environments, the ability to abstain when uncertain is crucial for patient safety. This metric addresses these limitations by quantifying the trade-off between accuracy and reliability.

Let $Q$ denote the complete set of all questions, $Q_\tau = \{q \in Q \mid c(q) \geq \tau\}$ represent questions with confidence at or above threshold $\tau$, and $Q_\tau^{\textit{correct}} = \{q \in Q_\tau \mid a(q) \text{ is correct and } c(q) \geq \tau\}$ contain correctly answered questions with confidence at or above $\tau$.

We define several metrics that vary with the confidence threshold $\tau$. The Hallucination Rate (HR) measures the proportion of incorrectly answered questions among attempted questions:

\begin{equation}
HR(\tau) = 1 - CA(\tau) = \frac{|Q_\tau| - |Q_\tau^{\textit{correct}}|}{|Q_\tau|}.
\end{equation}

where $CA(\tau) = |Q_\tau^{\textit{correct}}| / |Q_\tau|$ represents the Conditional Accuracy, the proportion of correct answers among questions with confidence above threshold $\tau$.

The Overall Accuracy (OA) captures the proportion of correctly answered questions among all questions in the dataset:

\begin{equation}
OA(\tau) = \frac{|Q_\tau^{\textit{correct}}|}{|Q|} = CA(\tau) \times RR(\tau).
\end{equation}

where $RR(\tau) = |Q_\tau| / |Q|$ is the Response Rate, representing the fraction of questions that the agent attempts to answer at threshold $\tau$.

Building on these definitions, we formulate HCAcc@k\% as the maximum overall accuracy achievable while maintaining a hallucination rate at or below $(100-k)\%$:

\begin{equation}
\label{eq:hcacc}
\max_{\tau} \{OA(\tau) \mid HR(\tau) \leq \frac{100-k}{100}\}.
\end{equation}

For example, HCAcc@90\% represents the maximum achievable accuracy when at most 10\% of provided answers can be incorrect. This metric provides a principled approach to evaluating confidence-based selective answering agents, effectively balancing utility against reliability.

\begin{table*}[t]
\centering
\small
\setlength{\tabcolsep}{4pt} 
\renewcommand{\arraystretch}{0.9}
\begin{tabular}{l|cccc|cccc}
\toprule
\multirow{2}{*}{\textbf{Method}} & \multicolumn{4}{c|}{\textbf{MIMIC-III}} & \multicolumn{4}{c}{\textbf{eICU}} \\
\cmidrule(lr){2-5} \cmidrule(lr){6-9}
& \textbf{@0\%} & \textbf{@50\%} & \textbf{@70\%} & \textbf{@90\%} & \textbf{@0\%} & \textbf{@50\%} & \textbf{@70\%} & \textbf{@90\%} \\
\midrule
\multicolumn{9}{l}{\textit{Without Confidence Estimation}} \\
\midrule
CoT \cite{wei2022chain} & 9.58 & \multicolumn{3}{c|}{\cellcolor{gray!15}} & 27.34 & \multicolumn{3}{c}{\cellcolor{gray!15}} \\
ReACT \cite{yao2023react} & 10.38 & \multicolumn{3}{c|}{\cellcolor{gray!15}} & 33.33 & \multicolumn{3}{c}{\cellcolor{gray!15}} \\
Reflexion \cite{shinn2023reflexion} & 19.48 & \multicolumn{3}{c|}{\cellcolor{gray!15}} & 36.72 & \multicolumn{3}{c}{\cellcolor{gray!15}} \\
DIN-SQL \cite{pourreza2023din} & 38.45 & \multicolumn{3}{c|}{\cellcolor{gray!15}N/A} & 25.00 & \multicolumn{3}{c}{\cellcolor{gray!15}N/A} \\
SQLAgent \cite{smolagents} & 52.49 & \multicolumn{3}{c|}{\cellcolor{gray!15}} & 37.41 & \multicolumn{3}{c}{\cellcolor{gray!15}} \\
PythonAgent \cite{smolagents} & 50.60 & \multicolumn{3}{c|}{\cellcolor{gray!15}} & 43.27 & \multicolumn{3}{c}{\cellcolor{gray!15}} \\
EHRAgent \cite{shi2024ehragent} & 58.97 & \multicolumn{3}{c|}{\cellcolor{gray!15}} & 53.10 & \multicolumn{3}{c}{\cellcolor{gray!15}} \\
\midrule
\multicolumn{9}{l}{\textit{With Confidence Estimation}} \\
\midrule
SQLAgent & 50.25 & 50.25 & 0.0 & 0.0 & 40.51 & 0.0 & 0.0 & 0.0 \\
PythonAgent & 48.02 & 44.06 & 0.0 & 0.0 & 40.51 & 0.0 & 0.0 & 0.0 \\
TrustEHRAgent & \textbf{62.13} & \textbf{62.13} & \textbf{44.23} & \textbf{3.27} & \textbf{59.82} & \textbf{59.82} & \textbf{25.34} & \textbf{2.24} \\
\bottomrule
\end{tabular}
\caption{Comparison of HCAcc@k\% scores across models and datasets. HCAcc@0\% corresponds to standard accuracy without accounting for hallucinations, while HCAcc@50\%–90\% reflect performance under increasingly strict reliability constraints.}
\label{tab:results}
\end{table*}

\section{Experimental Setup}
\label{sec:experiments}
\subsection{Dataset}
We evaluate our approach using a QA dataset \cite{shi2024ehragent} derived from EHRSQL \cite{lee2022ehrsql}, a Text-to-SQL benchmark built on MIMIC-III and eICU Electronic Health Records. The dataset contains 580 question-answer pairs for each database, ranging from simple retrievals to complex queries with calculations and multi-table joins. Dataset details are provided in Appendix~\ref{app:dataset}.

\subsection{Baseline Setup}

We conduct our evaluation from two perspectives to comprehensively assess TrustEHRAgent's performance. First, to evaluate reliability performance across different confidence thresholds, we compare the HCAcc@k\% curves for confidence-aware agents: our proposed TrustEHRAgent and baseline methods with confidence estimation including SQLAgent and PythonAgent variants \cite{smolagents}. This comparison specifically targets the agents' ability to appropriately abstain from answering when confidence is low, revealing how effectively each method can balance accuracy and reliability in clinical question answering scenarios.

Second, we verify that our confidence-aware architecture maintains competitive question answering performance by comparing its base accuracy (HCAcc@0\%) against established baselines including EHRAgent and six other methods previously evaluated in \cite{shi2024ehragent}. This determines whether our approach sacrifices fundamental QA capabilities while adding reliability mechanisms.
Implementation details for baseline are provided in Appendix~\ref{app:baseline}.

\subsection{Evaluation Metrics}
We adopt Hallucination-Controlled Accuracy at k\% (HCAcc@k\%) as our primary evaluation metric.
We report performance at multiple operating points to comprehensively evaluate agent behavior across varying reliability requirements. HCAcc@0\% serves as a baseline equivalent to traditional accuracy, measuring performance when the agent attempts all questions without considering confidence. HCAcc@50\% through HCAcc@90\% represent increasingly stringent reliability constraints that reflect clinical scenarios where patient safety demands high confidence in agent outputs.
We evaluate answer correctness using exact match, where the agent's response must precisely match the reference answer to be considered successful.


\section{Results}
\label{sec:results}

\subsection{Main Results}
Table~\ref{tab:results} demonstrates that TrustEHRAgent substantially outperforms baseline methods across MIMIC-III and eICU datasets, with the most pronounced improvements observed under strict reliability constraints. The reliability-constrained evaluation reveals the key strengths of our approach. At moderate reliability thresholds (HCAcc@50\%), TrustEHRAgent maintains its baseline accuracy performance on both datasets, while baseline methods already show performance degradation. As reliability requirements become more stringent, the performance gap widens dramatically. Under strict reliability constraints (HCAcc@70\%), TrustEHRAgent achieves notable accuracy improvements of 44.23\% on MIMIC-III and 25.34\% on eICU, while baseline methods fail to provide any predictions.

Under severe reliability constraints (HCAcc@90\%), all methods struggle considerably. While TrustEHRAgent still provides predictions with 3.27\% and 2.24\% accuracy on MIMIC-III and eICU respectively, this highlights that achieving high accuracy at extremely stringent reliability thresholds remains a considerable challenge and represents an important direction for future research in trustworthy clinical AI agents.

Additionally, our confidence-aware architecture preserves strong performance in traditional accuracy settings. Under HCAcc@0\%, TrustEHRAgent achieves 62.13\% and 59.82\% on MIMIC-III and eICU respectively, representing improvements of 3.16\% and 6.72\% over the original EHRAgent (58.97\%, 53.10\%). This confirms that adding reliability mechanisms does not compromise core question answering capabilities and even provides modest performance gains.

\begin{figure}[h]
    \centering
    \includegraphics[width=\linewidth]{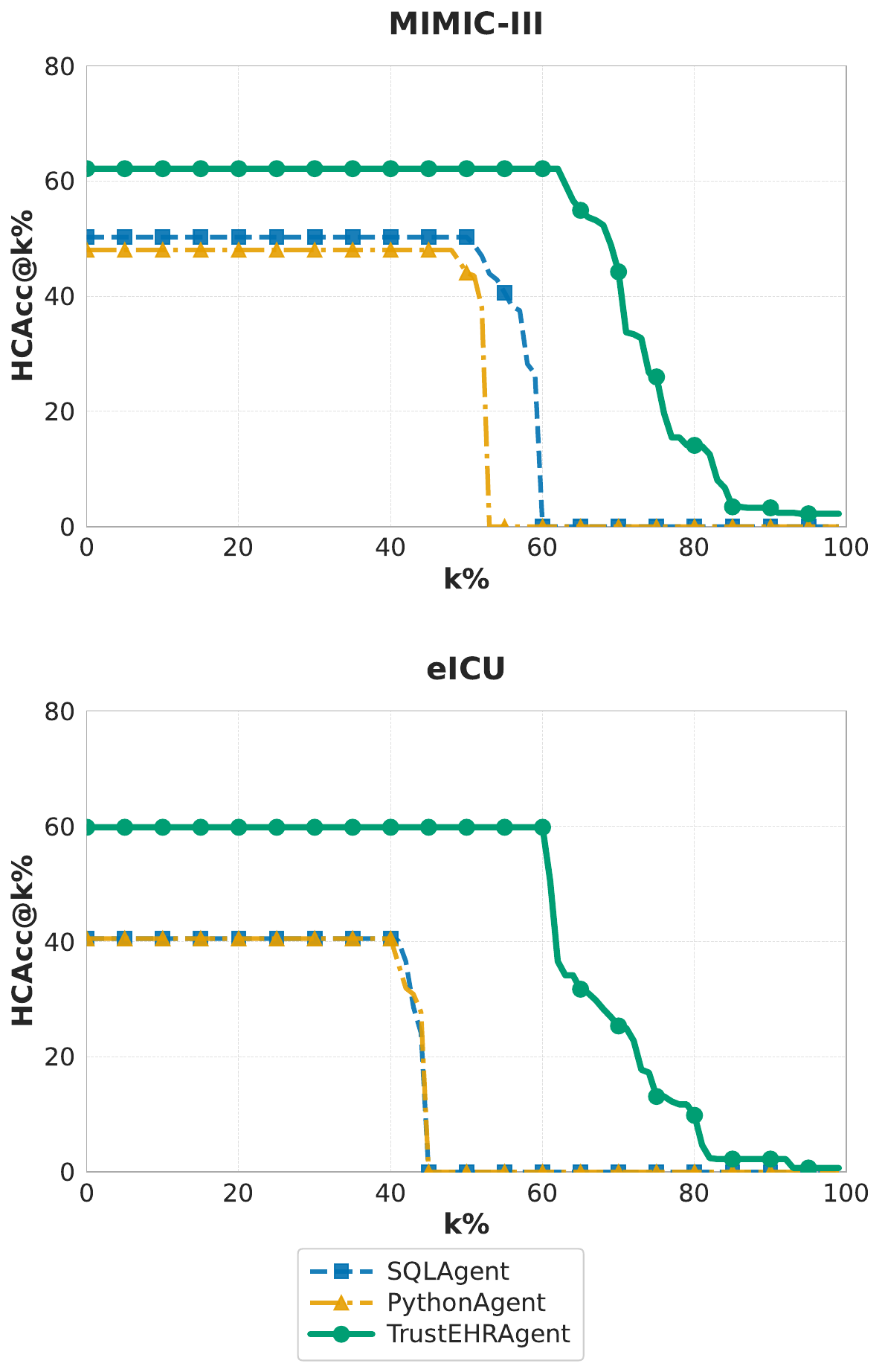}
    \caption{Performance-reliability trade-off analysis across models on MIMIC-III and eICU datasets. TrustEHRAgent maintains superior performance at high reliability thresholds compared to baseline methods.}
    \label{fig:performance_reliability}
\end{figure}

These results clearly demonstrate the necessity of evaluation frameworks that go beyond simple accuracy for medical AI agents. As shown in Figure~\ref{fig:performance_reliability}, models with minor differences in traditional accuracy metrics exhibit dramatically widening performance gaps as reliability constraints become more stringent. This provides compelling evidence for the importance of confidence estimation in clinical agents and validates the need for new evaluation paradigms like HCAcc@k\% that prioritize both performance and safety considerations.

\section{Analysis}

\subsection{Effectiveness of Step-wise Approach}
\begin{figure}[!h]
    \centering
    \includegraphics[width=\columnwidth]{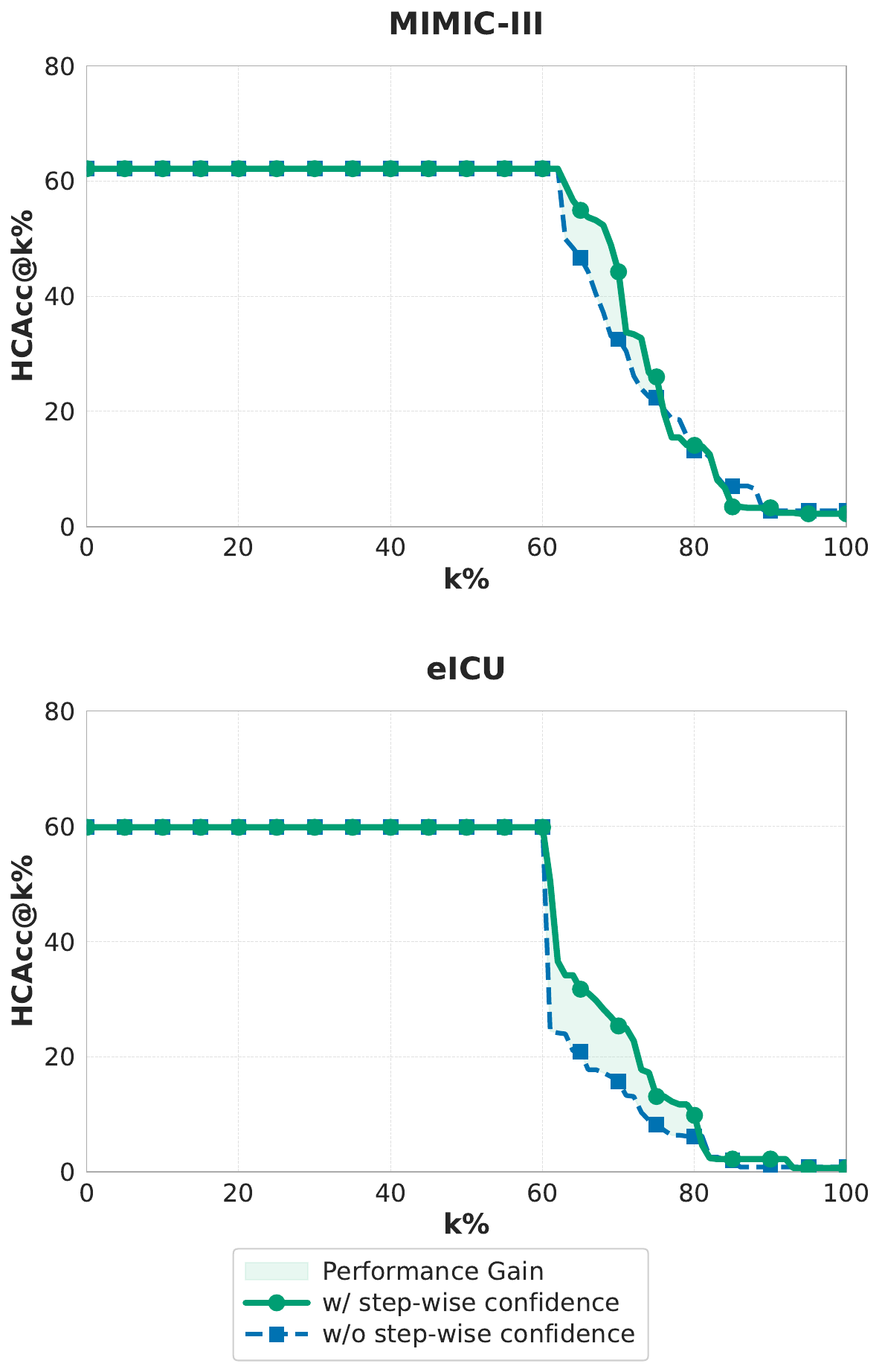}
    \caption{TrustEHRAgent with step-wise confidence estimation versus TrustEHRAgent without step-wise confidence estimation. Shaded areas show performance gains, with significant advantages at higher reliability thresholds.}
    \label{fig:confidence_ablation}
\end{figure}

To evaluate our step-wise confidence estimation approach, we compared TrustEHRAgent against a variant without the step-wise confidence estimation component. The key difference is whether step-wise verbalized confidence signals are included in the reasoning process. Figure~\ref{fig:confidence_ablation} shows the performance across MIMIC-III and eICU datasets. Differences emerge at higher thresholds where step-wise confidence estimation provides performance gains, particularly in the critical reliability range essential for clinical safety. This demonstrates that uncertainty information from each reasoning step significantly improves the accuracy of overall confidence estimation, confirming the utility of the step-wise approach in complex clinical reasoning.

\subsection{Comparison of Confidence Estimation Methods}
\begin{figure}[!t]
    \centering
    \includegraphics[width=\columnwidth]{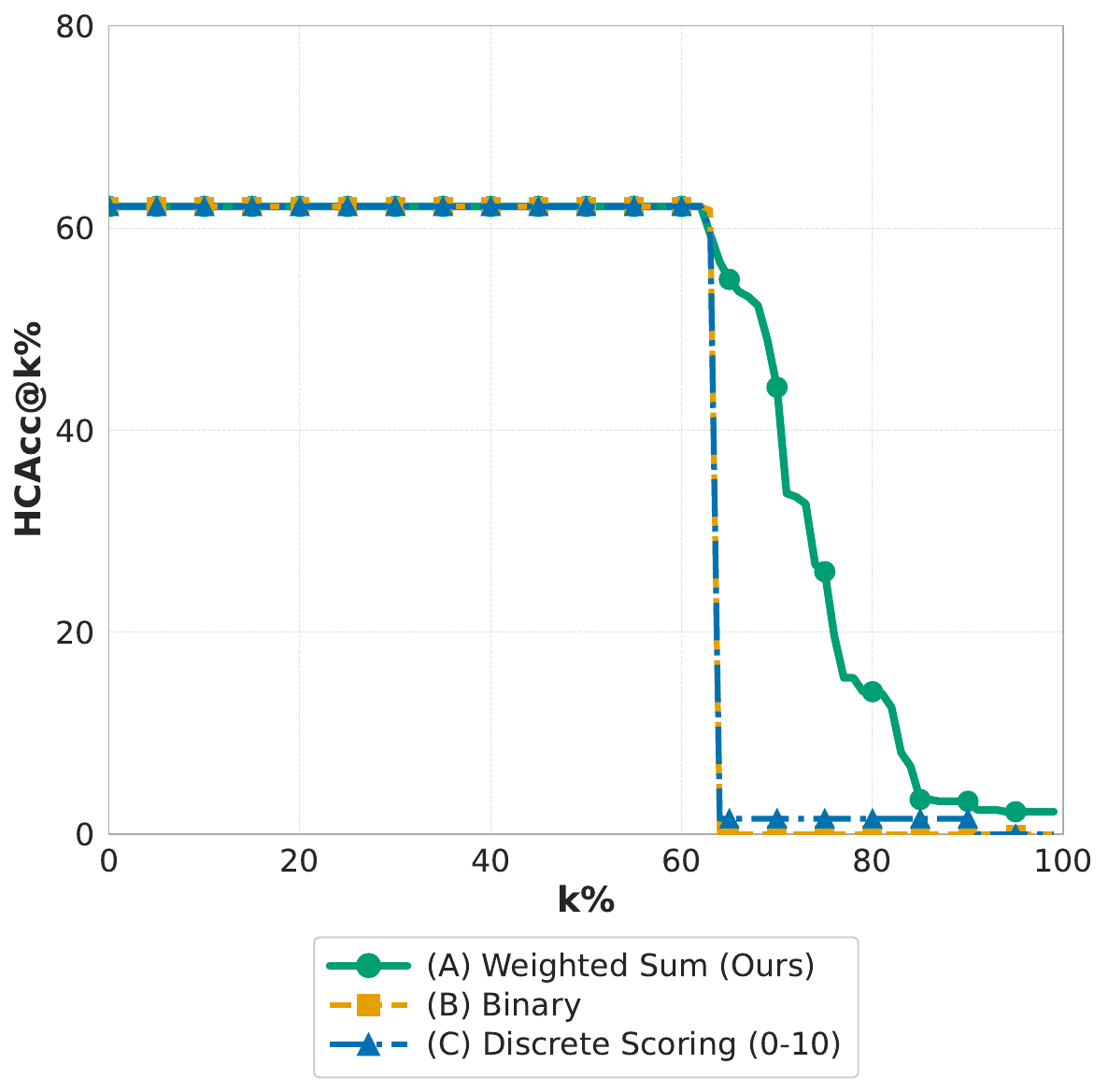}
    \caption{Comparison of confidence estimation approaches: (A) Our weighted sum approach based on log probabilities, (B) Binary classification, and (C) Discrete scoring.}
    \label{fig:confidence_methods}
\end{figure}

As illustrated in Figure~\ref{fig:confidence_methods}, we compare three distinct confidence estimation approaches. Our weighted sum approach (A) obtains a probability distribution over confidence levels using token log probabilities, enabling continuous confidence values. Binary classification approaches (B) use an LLM-based binary classifier to make simple accept/reject decisions about answer trustworthiness. Discrete scoring methods (C) assign numerical confidence scores from a fixed range (e.g., 0-10), providing limited discrete confidence levels.

The performance curves reveal substantial differences in their effectiveness. Our approach produces smooth performance curves due to its continuous confidence spectrum. Binary classification results in sharp performance drops as it cannot provide flexible k-threshold settings and fails to appropriately capture answers that should be rejected. Discrete scoring fails to provide effective confidence estimation because it restricts confidence values to a small set of predefined levels, lacking the nuanced distinctions needed for reliable filtering. This finding aligns with observations from \citet{liu2023g}, which demonstrated that probability-based continuous scoring outperforms discrete scoring methods by providing more fine-grained scores that better capture subtle differences between outputs.

\subsection{Case Study Analysis}

To illustrate the practical benefits of our confidence estimation approach, we examine two contrasting clinical scenarios from the MIMIC-III evaluation dataset, as detailed in Table~\ref{tab:confidence_cases}. These cases show how step-wise confidence estimation enables appropriate safety decisions in healthcare AI agents.

The first case demonstrates successful performance on a complex analytical query requiring identification of frequently prescribed medications following surgical heart complications. This multi-step process involves mapping diagnostic terminology to standardized ICD codes, performing temporal joins across multiple database tables, and aggregating prescription frequencies. Throughout this complex reasoning chain, TrustEHRAgent maintains consistently high confidence levels (0.95 $\rightarrow$ 0.92 $\rightarrow$ 0.90). The final confidence of 0.99 enables the system to provide the accurate medication list that matches the ground truth, supporting clinical decision-making.

\begin{table*}[t]
\centering
\small
\setlength{\tabcolsep}{8pt}
\renewcommand{\arraystretch}{1.2}
\begin{tabular}{p{0.18\textwidth}p{0.38\textwidth}p{0.35\textwidth}}
\toprule
\textbf{Attribute} & \textbf{Case 1: Clinical Analytics Query} & \textbf{Case 2: Medical Abbreviation Query} \\
\midrule
\textbf{Clinical Query} & 
What are the top 5 drugs prescribed after "surg compl-heart" diagnosis since 2103? & 
What does "ac embl internl jug vein" stand for? \\
\midrule
\textbf{Ground Truth} & 
potassium chloride, insulin, furosemide, d5w, ns (post-surgical medications) & 
Acute venous embolism and thrombosis of internal jugular veins (serious vascular condition) \\
\midrule
\textbf{Reasoning Process} & 
(1) Identify diagnosis codes \newline
(2) Join diagnoses with prescriptions table \newline
(3) Aggregate and rank drug frequencies & 
(1) Search medical dictionary \newline
(2) Query procedure codes \newline
(3) Attempt individual term matching \\
\midrule
\textbf{Predicted Answer} & 
potassium chloride, insulin, furosemide, d5w, ns & 
\textit{No result found} \\
\midrule
\textbf{Step-wise Confidence} & 
\textbf{0.95} $\rightarrow$ \textbf{0.92} $\rightarrow$ \textbf{0.90} & 
\textbf{0.90} $\rightarrow$ \textbf{0.60} $\rightarrow$ \textbf{0.40} \\
\midrule
\textbf{Final Confidence} & 
\textbf{0.99} & 
\textbf{0.499} \\
\midrule
\textbf{System Decision} & 
Answer provided with high confidence & 
Answer rejected due to uncertainty \\
\bottomrule
\end{tabular}
\caption{Case studies illustrating how step-wise confidence estimation enables appropriate clinical safety decisions in TrustEHRAgent.}
\label{tab:confidence_cases}
\end{table*}

In contrast, the second case involves interpreting an abbreviated medical term "ac embl internl jug vein," representing a common challenge in clinical settings where informal abbreviations may be used inconsistently across healthcare institutions. TrustEHRAgent initially approaches this query with moderate confidence (0.90) when searching the medical dictionary. However, as subsequent searches in procedure codes and individual term matching yield no definitive matches, the confidence systematically decreases (0.90 $\rightarrow$ 0.60 $\rightarrow$ 0.40). After completing the full reasoning process, the final confidence score of 0.499 indicates significant uncertainty. The system appropriately abstains from providing an answer, preventing potential misinterpretation of this medical abbreviation that could lead to clinical errors.

These contrasting confidence trajectories highlight the value of step-wise confidence monitoring for clinical AI safety. The consistently high confidence validates the reliability of complex analytical processes, while the declining confidence serves as an indicator of reasoning uncertainty. This demonstrates how confidence-aware systems can balance utility and safety in high-stakes medical environments by filtering out potentially unreliable responses before they reach clinicians.

\section{Conclusion}

Clinical AI agents encounter substantial reliability challenges in high-stakes healthcare environments where incorrect information can compromise patient safety. Traditional accuracy metrics inadequately assess these risks by assuming models should attempt every question regardless of uncertainty, failing to capture the critical reliability requirements essential for medical applications.

To address these limitations, we propose Hallucination-Controlled Accuracy at k\% (HCAcc@k\%), a novel evaluation metric that quantifies the accuracy-reliability trade-off by measuring maximum achievable accuracy under controlled hallucination constraints.  We also introduce TrustEHRAgent, an agent framework incorporating step-wise confidence estimation that exhibits superior performance while effectively balancing utility with safety requirements. 

Our findings validate the importance of reliability-aware evaluation in medical AI. While baseline methods achieved 0\% performance under stringent reliability constraints (HCAcc@70\%), TrustEHRAgent maintained substantial performance with 44.23\%p on MIMIC-III and 25.34\%p on eICU, illustrating its ability to provide accurate information when confident while appropriately abstaining when uncertain.

This work represents an important advancement toward developing trustworthy AI agents that safely assist healthcare professionals in interpreting EHR data. Our approach establishes a foundation for safe deployment of AI agents in high-risk clinical environments while prioritizing patient safety.

\section*{Limitations}

\textbf{Performance at High Reliability Thresholds} Despite significant improvements over baselines, TrustEHRAgent still achieves relatively low accuracy (3.27\% on MIMIC-III and 2.24\% on eICU) at the most stringent reliability threshold (HCAcc@90\%). This indicates that developing agents capable of maintaining high accuracy under extreme reliability constraints remains challenging and warrants further research.

\textbf{Dataset Limitations} Our evaluation relies on MIMIC-III and eICU datasets which, while extensive, may not fully represent the diversity of real-world clinical settings, patient populations, or EHR systems. The 580 question-answer pairs per database, though comprehensive, might not capture the full spectrum of clinical queries encountered in practice.

\textbf{Need for Advanced Calibration Methods} Our current weighted log probability approach, while effective, could benefit from more sophisticated calibration techniques. Future work should explore methods such as temperature scaling, ensemble approaches, or Bayesian calibration to further improve the alignment between confidence scores and actual performance, particularly at higher reliability thresholds. Developing domain-specific calibration methods tailored to clinical contexts could significantly enhance the practical utility of confidence-aware agents in healthcare environments.

\section*{Ethics Statement}
This study utilizes the MIMIC-III and eICU datasets, which contain de-identified electronic health records with rigorous privacy protections. Access to these datasets was obtained following required institutional protocols, including completion of the CITI "Data or Specimens Only Research" course and Data Use Agreement for MIMIC-III, and the PhysioNet Credentialed Health Data Use Agreement for eICU. No protected health information (PHI) was accessed, as both datasets have undergone thorough de-identification processes in compliance with HIPAA Safe Harbor provisions. All experimental procedures were conducted in accordance with relevant institutional regulations for secondary analysis of de-identified health data. No attempt was made to re-identify patients, and all results are presented in aggregate form to preserve anonymity. AI assistance tools were used solely for writing assistance during manuscript preparation.

\bibliography{custom}

\begin{thebibliography}{30}
\providecommand{\natexlab}[1]{#1}

\bibitem[{Agarwal et~al.(2024)Agarwal, Jin, Chandra, De~Choudhury, Kumar, and Sastry}]{agarwal2024medhalu}
Vibhor Agarwal, Yiqiao Jin, Mohit Chandra, Munmun De~Choudhury, Srijan Kumar, and Nishanth Sastry. 2024.
\newblock Medhalu: Hallucinations in responses to healthcare queries by large language models.
\newblock \emph{arXiv preprint arXiv:2409.19492}.

\bibitem[{Ahmad et~al.(2023)Ahmad, Yaramis, and Roy}]{ahmad2023creating}
Muhammad~Aurangzeb Ahmad, Ilker Yaramis, and Taposh~Dutta Roy. 2023.
\newblock Creating trustworthy llms: Dealing with hallucinations in healthcare ai.
\newblock \emph{arXiv preprint arXiv:2311.01463}.

\bibitem[{Ahsan et~al.(2024)Ahsan, McInerney, Kim, Potter, Young, Amir, and Wallace}]{ahsan2024retrieving}
Hiba Ahsan, Denis~Jered McInerney, Jisoo Kim, Christopher Potter, Geoffrey Young, Silvio Amir, and Byron~C Wallace. 2024.
\newblock Retrieving evidence from ehrs with llms: possibilities and challenges.
\newblock \emph{Proceedings of machine learning research}, 248:489.

\bibitem[{Chen and Mueller(2023)}]{jiuhai2023}
Jiuhai Chen and Jonas Mueller. 2023.
\newblock Quantifying uncertainty in answers from any language model and enhancing their trustworthiness.
\newblock \emph{arXiv (Cornell University)}.

\bibitem[{Gekhman et~al.(2025)Gekhman, David, Orgad, Ofek, Belinkov, Szpektor, Herzig, and Reichart}]{gekhman2025inside}
Zorik Gekhman, Eyal~Ben David, Hadas Orgad, Eran Ofek, Yonatan Belinkov, Idan Szpektor, Jonathan Herzig, and Roi Reichart. 2025.
\newblock Inside-out: Hidden factual knowledge in llms.
\newblock \emph{arXiv preprint arXiv:2503.15299}.

\bibitem[{Hager et~al.(2024)Hager, Jungmann, Holland, Bhagat, Hubrecht, Knauer, Vielhauer, Makowski, Braren, Kaissis et~al.}]{hager2024evaluation}
Paul Hager, Friederike Jungmann, Robbie Holland, Kunal Bhagat, Inga Hubrecht, Manuel Knauer, Jakob Vielhauer, Marcus Makowski, Rickmer Braren, Georgios Kaissis, and 1 others. 2024.
\newblock Evaluation and mitigation of the limitations of large language models in clinical decision-making.
\newblock \emph{Nature medicine}, 30(9):2613--2622.

\bibitem[{Han et~al.(2024)Han, Buntine, and Shareghi}]{han2024towards}
Jiuzhou Han, Wray Buntine, and Ehsan Shareghi. 2024.
\newblock Towards uncertainty-aware language agent.
\newblock \emph{arXiv preprint arXiv:2401.14016}.

\bibitem[{Jiang et~al.(2020)Jiang, Araki, Ding, and Neubig}]{zhengbao2020}
Zhengbao Jiang, Jun Araki, Haibo Ding, and Graham Neubig. 2020.
\newblock How can we know when language models know? on the calibration of language models for question answering.
\newblock \emph{arXiv (Cornell University)}.

\bibitem[{Lee et~al.(2022)Lee, Hwang, Bae, Kwon, Shin, Yang, Seo, Kim, and Choi}]{lee2022ehrsql}
Gyubok Lee, Hyeonji Hwang, Seongsu Bae, Yeonsu Kwon, Woncheol Shin, Seongjun Yang, Minjoon Seo, Jong-Yeup Kim, and Edward Choi. 2022.
\newblock Ehrsql: A practical text-to-sql benchmark for electronic health records.
\newblock \emph{Advances in Neural Information Processing Systems}, 35:15589--15601.

\bibitem[{Li et~al.(2023)Li, Cheng, Zhao, Nie, and Wen}]{li2023halueval}
Junyi Li, Xiaoxue Cheng, Wayne~Xin Zhao, Jian-Yun Nie, and Ji-Rong Wen. 2023.
\newblock Halueval: A large-scale hallucination evaluation benchmark for large language models.
\newblock In \emph{Proceedings of the 2023 Conference on Empirical Methods in Natural Language Processing}, pages 6449--6464.

\bibitem[{Lin et~al.(2021)Lin, Hilton, and Evans}]{lin2021truthfulqa}
Stephanie Lin, Jacob Hilton, and Owain Evans. 2021.
\newblock Truthfulqa: Measuring how models mimic human falsehoods, 2022.
\newblock \emph{URL https://arxiv. org/abs/2109.07958}.

\bibitem[{Lin et~al.(2023)Lin, Trivedi, and Sun}]{lin2023generating}
Zhen Lin, Shubhendu Trivedi, and Jimeng Sun. 2023.
\newblock Generating with confidence: Uncertainty quantification for black-box large language models.
\newblock \emph{arXiv preprint arXiv:2305.19187}.

\bibitem[{Liu et~al.(2023)Liu, Iter, Xu, Wang, Xu, and Zhu}]{liu2023g}
Yang Liu, Dan Iter, Yichong Xu, Shuohang Wang, Ruochen Xu, and Chenguang Zhu. 2023.
\newblock G-eval: Nlg evaluation using gpt-4 with better human alignment.
\newblock \emph{arXiv preprint arXiv:2303.16634}.

\bibitem[{Mahaut et~al.(2024)Mahaut, Aina, Czarnowska, Hardalov, Mueller, and M{\`a}rquez}]{mahaut2024factual}
Mat{\'e}o Mahaut, Laura Aina, Paula Czarnowska, Momchil Hardalov, Thomas Mueller, and Llu{\'\i}s M{\`a}rquez. 2024.
\newblock Factual confidence of llms: on reliability and robustness of current estimators.
\newblock In \emph{Proceedings of the 62nd Annual Meeting of the Association for Computational Linguistics (Volume 1: Long Papers)}, pages 4554--4570.

\bibitem[{Mehenni and Zouaq(2025)}]{mehenni2025medhal}
Gaya Mehenni and Amal Zouaq. 2025.
\newblock Medhal: An evaluation dataset for medical hallucination detection.
\newblock \emph{arXiv preprint arXiv:2504.08596}.

\bibitem[{Orgad et~al.(2024)Orgad, Toker, Gekhman, Reichart, Szpektor, Kotek, and Belinkov}]{orgad2024llms}
Hadas Orgad, Michael Toker, Zorik Gekhman, Roi Reichart, Idan Szpektor, Hadas Kotek, and Yonatan Belinkov. 2024.
\newblock Llms know more than they show: On the intrinsic representation of llm hallucinations.
\newblock \emph{arXiv preprint arXiv:2410.02707}.

\bibitem[{Pal et~al.(2023)Pal, Umapathi, and Sankarasubbu}]{pal2023med}
Ankit Pal, Logesh~Kumar Umapathi, and Malaikannan Sankarasubbu. 2023.
\newblock Med-halt: Medical domain hallucination test for large language models.
\newblock In \emph{Proceedings of the 27th Conference on Computational Natural Language Learning (CoNLL)}, pages 314--334.

\bibitem[{Pourreza and Rafiei(2023)}]{pourreza2023din}
Mohammadreza Pourreza and Davood Rafiei. 2023.
\newblock Din-sql: Decomposed in-context learning of text-to-sql with self-correction.
\newblock \emph{Advances in Neural Information Processing Systems}, 36:36339--36348.

\bibitem[{Qiu et~al.(2024)Qiu, Lam, Li, Acharya, Wong, Darzi, Yuan, and Topol}]{qiu2024llm}
Jianing Qiu, Kyle Lam, Guohao Li, Amish Acharya, Tien~Yin Wong, Ara Darzi, Wu~Yuan, and Eric~J Topol. 2024.
\newblock Llm-based agentic systems in medicine and healthcare.
\newblock \emph{Nature Machine Intelligence}, 6(12):1418--1420.

\bibitem[{Roucher et~al.(2025)Roucher, del Moral, Wolf, von Werra, and Kaunismäki}]{smolagents}
Aymeric Roucher, Albert~Villanova del Moral, Thomas Wolf, Leandro von Werra, and Erik Kaunismäki. 2025.
\newblock `smolagents`: a smol library to build great agentic systems.
\newblock \url{https://github.com/huggingface/smolagents}.

\bibitem[{Shi et~al.(2024)Shi, Xu, Zhuang, Yu, Zhang, Wu, Zhu, Ho, Yang, and Wang}]{shi2024ehragent}
Wenqi Shi, Ran Xu, Yuchen Zhuang, Yue Yu, Jieyu Zhang, Hang Wu, Yuanda Zhu, Joyce Ho, Carl Yang, and May~Dongmei Wang. 2024.
\newblock Ehragent: Code empowers large language models for few-shot complex tabular reasoning on electronic health records.
\newblock In \emph{Proceedings of the 2024 Conference on Empirical Methods in Natural Language Processing}, pages 22315--22339.

\bibitem[{Shinn et~al.(2023)Shinn, Cassano, Labash, Gopinath, Narasimhan, and Yao}]{shinn2023reflexion}
Noah Shinn, Federico Cassano, Beck Labash, Ashwin Gopinath, Karthik Narasimhan, and Shunyu Yao. 2023.
\newblock Reflexion: Language agents with verbal reinforcement learning, 2023.
\newblock \emph{URL https://arxiv. org/abs/2303.11366}.

\bibitem[{Wang et~al.(2020)Wang, Shi, and Reddy}]{wang2020text}
Ping Wang, Tian Shi, and Chandan~K Reddy. 2020.
\newblock Text-to-sql generation for question answering on electronic medical records.
\newblock In \emph{Proceedings of The Web Conference 2020}, pages 350--361.

\bibitem[{Wang et~al.(2023)Wang, Haddow, Birch, and Peng}]{wang2023assessing}
Weixuan Wang, Barry Haddow, Alexandra Birch, and Wei Peng. 2023.
\newblock Assessing the reliability of large language model knowledge.
\newblock \emph{arXiv preprint arXiv:2310.09820}.

\bibitem[{Wei et~al.(2022)Wei, Wang, Schuurmans, Bosma, Xia, Chi, Le, Zhou et~al.}]{wei2022chain}
Jason Wei, Xuezhi Wang, Dale Schuurmans, Maarten Bosma, Fei Xia, Ed~Chi, Quoc~V Le, Denny Zhou, and 1 others. 2022.
\newblock Chain-of-thought prompting elicits reasoning in large language models.
\newblock \emph{Advances in neural information processing systems}, 35:24824--24837.

\bibitem[{Wu et~al.(2023)Wu, Lin, Fang, and Huang}]{wu2023medical}
Chengyan Wu, Zehong Lin, Wenlong Fang, and Yuyan Huang. 2023.
\newblock A medical diagnostic assistant based on llm.
\newblock In \emph{China Health Information Processing Conference}, pages 135--147. Springer.

\bibitem[{Xiong et~al.(2023)Xiong, Hu, Lu, Li, Fu, He, and Hooi}]{xiong2023can}
Miao Xiong, Zhiyuan Hu, Xinyang Lu, Yifei Li, Jie Fu, Junxian He, and Bryan Hooi. 2023.
\newblock Can llms express their uncertainty? an empirical evaluation of confidence elicitation in llms.
\newblock \emph{arXiv preprint arXiv:2306.13063}.

\bibitem[{Yang et~al.(2024)Yang, Tsai, and Yamada}]{yang2024verbalizedconfidencescoresllms}
Daniel Yang, Yao-Hung~Hubert Tsai, and Makoto Yamada. 2024.
\newblock \href {https://arxiv.org/abs/2412.14737} {On verbalized confidence scores for llms}.
\newblock \emph{Preprint}, arXiv:2412.14737.

\bibitem[{Yang et~al.(2022)Yang, Chen, PourNejatian, Shin, Smith, Parisien, Compas, Martin, Costa, Flores et~al.}]{yang2022large}
Xi~Yang, Aokun Chen, Nima PourNejatian, Hoo~Chang Shin, Kaleb~E Smith, Christopher Parisien, Colin Compas, Cheryl Martin, Anthony~B Costa, Mona~G Flores, and 1 others. 2022.
\newblock A large language model for electronic health records.
\newblock \emph{NPJ digital medicine}, 5(1):194.

\bibitem[{Yao et~al.(2023)Yao, Zhao, Yu, Du, Shafran, Narasimhan, and Cao}]{yao2023react}
Shunyu Yao, Jeffrey Zhao, Dian Yu, Nan Du, Izhak Shafran, Karthik Narasimhan, and Yuan Cao. 2023.
\newblock React: Synergizing reasoning and acting in language models.
\newblock In \emph{International Conference on Learning Representations (ICLR)}.

\end{thebibliography}

\newpage

\appendix

\section{Step-wise Confidence Estimation}
\label{app:step-wise}

Step-wise confidence estimation enables TrustEHRAgent to track uncertainty throughout the reasoning process by providing explicit confidence ratings at each major reasoning step. This component is implemented through prompt-based instructions integrated into the agent's reasoning workflow. The agent is instructed to provide explicit confidence ratings during code generation and database interaction steps using the following prompt addition: "For each code solution, include a confidence rating on a scale from 0-10 at the top of your code as a comment in the format '\# Confidence: X' (where X is a number from 0 to 10). Use a highly critical standard for this rating. Consider data limitations, interpretation ambiguities, and error propagation risks. Use higher scores only when genuinely confident." This instruction ensures that the agent explicitly evaluates its uncertainty at each reasoning step, particularly during SQL query generation and data interpretation phases where errors are most likely to occur and propagate through the reasoning chain. The step-wise confidence scores are collected throughout the reasoning process and incorporated as contextual information for the final confidence estimation described in Section~\ref{sec:trustehragent}, providing valuable signals about reasoning quality and uncertainty evolution that enhance the reliability of threshold-based decision making.

\section{Confidence Estimator Implementation Details}
\label{app:confidence}

This appendix provides the prompt template and implementation details for the Confidence Estimator component in TrustEHRAgent. The Confidence Estimator takes three inputs: the original clinical question, the agent's reasoning history with step-wise confidence signals, and the proposed final answer. It then outputs a confidence score using the weighted sum approach described in Section~\ref{sec:trustehragent}. The confidence estimator uses the same GPT-4.1-mini model as the agent backbone to ensure consistency in language understanding and evaluation capabilities.

Table~\ref{tab:confidence_prompt} presents the structured prompt template used for confidence evaluation.


\section{Dataset Details}
\label{app:dataset}

We utilize the EHR database-based question-answer dataset proposed by EHRAgent \cite{shi2024ehragent}, which builds upon the foundation established by EHRSQL \cite{lee2022ehrsql}. This dataset represents a significant advancement in clinical question answering evaluation as it consists of real-world questions collected from 200 hospital staff members, ensuring authentic clinical information needs and realistic query complexity. The dataset comprehensively covers diverse question topics including demographics, prescriptions, laboratory results, and diagnostic information, while also incorporating time-sensitive questions that reflect the temporal nature of clinical decision-making.  

Our evaluation encompasses a total of 1,160 question-answer pairs, with 580 questions each for MIMIC-III and eICU databases. This balanced distribution enables comprehensive assessment across different clinical settings while maintaining sufficient statistical power for reliable performance evaluation. Each question-answer pair includes the original clinical question, the expected answer, and a gold standard SQL query that serves as the reference implementation for database interaction. Table~\ref{tab:mimic_examples} and Table~\ref{tab:eicu_examples} show examples of questions, answers, and corresponding gold SQL queries for MIMIC-III and eICU datasets, respectively.

\section{Baseline Setup}
\label{app:baseline}

We implement and evaluate our TrustEHRAgent against several established baselines from the EHRAgent paper \cite{shi2024ehragent}, ensuring fair comparison by using identical experimental conditions. All methods share the same (1) EHR metadata, (2) tool definitions, and (3) initial few-shot demonstrations in the prompts by default. The only differences lie in the prompting strategies and technical implementations of each method.

\subsection{Confidence-Aware Baselines}

For confidence estimation evaluation, we implement confidence-aware versions of the following baselines:

SQLAgent \cite{smolagents}: An SQL-based agent that generates SQL queries for database interactions. We augment this baseline with our confidence estimation framework by applying the same step-wise confidence prompting and final confidence estimator described in Section~\ref{sec:trustehragent}. The agent generates SQL commands to interact with EHR databases and provides confidence scores for each generated query.

PythonAgent \cite{smolagents}: A Python-based coding agent that generates Python scripts for EHR question answering. Similar to SQLAgent, we enhance this baseline with confidence estimation capabilities, enabling it to provide confidence scores alongside its Python-based solutions. The agent utilizes the same tool definitions and can execute Python code for complex reasoning tasks.

Both confidence-aware baselines follow the same confidence estimation pipeline as TrustEHRAgent, including step-wise confidence rating and the final confidence estimator using token log probabilities. This ensures that performance differences can be attributed to the underlying reasoning and code generation capabilities rather than the confidence estimation mechanism itself.

\subsection{Baselines without Confidence Estimation}

For comprehensive evaluation, we also compare against established methods without confidence estimation from the EHRAgent \cite{shi2024ehragent}. These include CoT \cite{wei2022chain}, which enhances reasoning by generating intermediate steps in natural language, and ReACT \cite{yao2023react}, a reasoning and acting framework that combines reasoning traces with tool usage for problem-solving. We also compare against Reflexion \cite{shinn2023reflexion}, a self-reflection approach that uses verbal reinforcement learning to improve performance through linguistic feedback, and DIN-SQL \cite{pourreza2023din}, a decomposed in-context learning approach for text-to-SQL generation with self-correction mechanisms. Additionally, we compare with the original SQLAgent \cite{smolagents} for SQL-based database query generation and PythonAgent \cite{smolagents} for Python-based coding tasks, both implemented without confidence estimation capabilities. Finally, we evaluate against EHRAgent, the state-of-the-art LLM agent for EHR question answering, which serves as our primary baseline for traditional accuracy comparison.

\subsection{Implementation Details}

All experiments use GPT-4.1-mini as the base language model with temperature set to 0.0 for deterministic outputs. We maintain consistent experimental conditions across all baselines, using identical EHR metadata and database schemas, tool function definitions and API descriptions, initial four-shot demonstrations (K=4), maximum interaction steps (T=10) for interactive methods, and evaluation criteria and answer matching procedures.

For confidence-aware baselines, we implement the same confidence estimation components as TrustEHRAgent, including step-wise confidence prompting with 0-10 scale ratings and the final confidence estimator using weighted token log probabilities over 5 discrete confidence levels (0-4).

\begin{table*}[h]
\centering
\begin{tabular}{|p{0.95\textwidth}|}
\hline
\vspace{0.3cm}

You are an expert evaluator assessing the reliability of a medical database query assistant's responses. Your task is to rate the assistant's answer on confidence based on the question, the solution process, and the final answer provided.

Confidence (0-4) - How confident should we be in the assistant's answer, where:
\begin{itemize}
\item 0: No confidence (completely incorrect or fundamentally flawed solution)
\item 1: Low confidence (partially correct but contains significant errors)
\item 2: Medium confidence (mostly correct but some uncertainty exists)
\item 3: High confidence (accurate and supported by appropriate SQL queries)
\item 4: Very high confidence (perfectly accurate with comprehensive SQL queries)
\end{itemize}

Evaluation Steps:
\begin{enumerate}
\item Examine the question to understand what medical database information is being requested.
\item Review the solution process, including the SQL queries performed, their results, and the assistant's own confidence rating.
\item Analyze the final answer to determine if:
\begin{enumerate}
\item It directly addresses the user's question
\item The SQL queries are correct and appropriate for the question
\item The reasoning process is sound and follows logical steps
\item The final calculation or extraction of data is accurate
\item The answer format matches what was requested in the question
\end{enumerate}
\item Assign a confidence score from 0-4 based on the evaluation criteria.
\end{enumerate}

\vspace{0.3cm} \\
\hline
\end{tabular}
\caption{Confidence Estimation Prompt Template}
\label{tab:confidence_prompt}
\end{table*}

\begin{table*}[t]
\centering
\small
\setlength{\tabcolsep}{4pt}
\renewcommand{\arraystretch}{1.1}
\begin{tabular}{p{0.25\textwidth}p{0.5\textwidth}p{0.2\textwidth}}
\toprule
\textbf{Question} & \textbf{Gold SQL Query} & \textbf{Answer} \\
\midrule
What is lidocaine 5\% ointment's way of ingesting it? & 
\texttt{SELECT DISTINCT prescriptions.route FROM prescriptions WHERE prescriptions.drug = 'lidocaine 5\% ointment'} & 
tp \\
\midrule
What is the last measured height of patient 26772 during the last hospital encounter? & 
\begin{minipage}[t]{0.5\textwidth}
\texttt{SELECT chartevents.valuenum FROM chartevents WHERE chartevents.icustay\_id IN (SELECT icustays.icustay\_id FROM icustays WHERE icustays.hadm\_id IN (SELECT admissions.hadm\_id FROM admissions WHERE admissions.subject\_id = 26772 AND admissions.dischtime IS NOT NULL ORDER BY admissions.admittime DESC LIMIT 1)) AND chartevents.itemid IN (SELECT d\_items.itemid FROM d\_items WHERE d\_items.label = 'admit ht' AND d\_items.linksto = 'chartevents') ORDER BY chartevents.charttime DESC LIMIT 1}
\end{minipage} & 
66.0 \\
\midrule
Is the value of patient 1205's last arterial bp [systolic] measured on the first intensive care unit visit less than the second to last value measured on the first intensive care unit visit? & 
\begin{minipage}[t]{0.5\textwidth}
\texttt{SELECT (SELECT chartevents.valuenum FROM chartevents WHERE chartevents.icustay\_id IN (SELECT icustays.icustay\_id FROM icustays WHERE icustays.hadm\_id IN (SELECT admissions.hadm\_id FROM admissions WHERE admissions.subject\_id = 1205) AND icustays.outtime IS NOT NULL ORDER BY icustays.intime ASC LIMIT 1) AND chartevents.itemid IN (SELECT d\_items.itemid FROM d\_items WHERE d\_items.label = 'arterial bp [systolic]' AND d\_items.linksto = 'chartevents') ORDER BY chartevents.charttime DESC LIMIT 1) < (SELECT chartevents.valuenum FROM chartevents WHERE chartevents.icustay\_id IN (SELECT icustays.icustay\_id FROM icustays WHERE icustays.hadm\_id IN (SELECT admissions.hadm\_id FROM admissions WHERE admissions.subject\_id = 1205) AND icustays.outtime IS NOT NULL ORDER BY icustays.intime ASC LIMIT 1) AND chartevents.itemid IN (SELECT d\_items.itemid FROM d\_items WHERE d\_items.label = 'arterial bp [systolic]' AND d\_items.linksto = 'chartevents') ORDER BY chartevents.charttime DESC LIMIT 1 OFFSET 1)}
\end{minipage} & 
1 \\
\bottomrule
\end{tabular}
\caption{Examples from the MIMIC-III evaluation dataset.}
\label{tab:mimic_examples}
\end{table*}

\begin{table*}[t]
\centering
\small
\setlength{\tabcolsep}{4pt}
\renewcommand{\arraystretch}{1.1}
\begin{tabular}{p{0.25\textwidth}p{0.5\textwidth}p{0.2\textwidth}}
\toprule
\textbf{Question} & \textbf{Gold SQL Query} & \textbf{Answer} \\
\midrule
What's the price for the antifungal therapy - caspofungin? & 
\texttt{SELECT DISTINCT cost.cost FROM cost WHERE cost.eventtype = 'treatment' AND cost.eventid IN (SELECT treatment.treatmentid FROM treatment WHERE treatment.treatmentname = 'antifungal therapy - caspofungin')} & 
16.26 \\
\midrule
In 2104, what were the top four most common laboratory tests that were ordered to patients in the same hospital encounter after being diagnosed with intracranial injury? & 
\begin{minipage}[t]{0.5\textwidth}
\texttt{SELECT t3.labname FROM (SELECT t2.labname, dense\_rank() OVER (ORDER BY count(*) DESC) AS c1 FROM (SELECT patient.uniquepid, diagnosis.diagnosistime, patient.patienthealthsystemstayid FROM diagnosis JOIN patient ON diagnosis.patientunitstayid = patient.patientunitstayid WHERE diagnosis.diagnosisname = 'intracranial injury' AND strftime('\%Y',diagnosis.diagnosistime) = '2104') AS t1 JOIN (SELECT patient.uniquepid, lab.labname, lab.labresulttime, patient.patienthealthsystemstayid FROM lab JOIN patient ON lab.patientunitstayid = patient.patientunitstayid WHERE strftime('\%Y',lab.labresulttime) = '2104') AS t2 ON t1.uniquepid = t2.uniquepid WHERE t1.diagnosistime < t2.labresulttime AND t1.patienthealthsystemstayid = t2.patienthealthsystemstayid GROUP BY t2.labname) AS t3 WHERE t3.c1 <= 4}
\end{minipage} & 
\begin{minipage}[t]{0.2\textwidth}
bedside glucose, hct, glucose, hgb, chloride
\end{minipage} \\
\midrule
Did patient 015-21757 receive any procedure since 4 years ago? & 
\begin{minipage}[t]{0.5\textwidth}
\texttt{SELECT count(*) > 0 FROM treatment WHERE treatment.patientunitstayid IN (SELECT patient.patientunitstayid FROM patient WHERE patient.patienthealthsystemstayid IN (SELECT patient.patienthealthsystemstayid FROM patient WHERE patient.uniquepid = '015-21757')) AND datetime(treatment.treatmenttime) >= datetime(current\_time, '-4 year')}
\end{minipage} & 
1 \\
\bottomrule
\end{tabular}
\caption{Examples from the eICU evaluation dataset.}
\label{tab:eicu_examples}
\end{table*}

\end{document}